\documentclass[sigconf]{acmart}

\usepackage{multirow}
\usepackage{tikz}
\usepackage{pgfplots}
\pgfplotsset{compat=1.16}

\AtBeginDocument{%
  \providecommand\BibTeX{{%
    \normalfont B\kern-0.5em{\scshape i\kern-0.25em b}\kern-0.8em\TeX}}}

\copyrightyear{2020}
\acmYear{2020}
\setcopyright{rightsretained}
\acmConference[ICAIF '20]{ACM International Conference on AI in Finance}{October 15--16, 2020}{New York, NY, USA}
\acmBooktitle{ACM International Conference on AI in Finance (ICAIF '20),October 15--16, 2020, New York, NY, USA}
\acmDOI{10.1145/3383455.3422549}
\acmISBN{978-1-4503-7584-9/20/10}

\begin{document}

\title[Detecting money laundering in the presence of label scarcity]{Machine learning methods to detect money laundering in the Bitcoin blockchain in the presence of label scarcity}

\author{Joana Lorenz}
\email{joana.lorenz@outlook.de}
\affiliation{\institution{NOVA IMS \& Feedzai}}

\author{Maria In\^es Silva}
\email{maria.silva@feedzai.com}
\affiliation{\institution{Feedzai}}

\author{David Apar\'icio}
\email{david.aparicio@feedzai.com}
\affiliation{\institution{Feedzai}}

\author{Jo\~ao Tiago Ascens\~ao}
\email{joao.ascensao@feedzai.com}
\affiliation{\institution{Feedzai}}

\author{Pedro Bizarro}
\email{pedro.bizarro@feedzai.com}
\affiliation{\institution{Feedzai}}

\renewcommand{\shortauthors}{Joana Lorenz et al.}

\begin{abstract}
Every year, criminals launder billions of dollars acquired from serious felonies (e.g., terrorism, drug smuggling, or human trafficking), harming countless people and economies. Cryptocurrencies, in particular, have developed as a haven for money laundering activity. 
Machine Learning can be used to detect these illicit patterns. However, labels are so scarce that traditional supervised algorithms are inapplicable. Here, we address money laundering detection assuming minimal access to labels. First, we show that existing state-of-the-art solutions using unsupervised anomaly detection methods are inadequate to detect the illicit patterns in a real Bitcoin transaction dataset. Then, we show that our proposed active learning solution is capable of matching the performance of a fully supervised baseline by using just 5\% of the labels. This solution mimics a typical real-life situation in which a limited number of labels can be acquired through manual annotation by experts.
\end{abstract}

\ccsdesc[300]{Computing methodologies~Supervised learning by classification}
\ccsdesc[300]{Computing methodologies~Anomaly detection}
\ccsdesc[300]{Computing methodologies~Active learning settings}
\ccsdesc[500]{Applied computing~Economics}

\keywords{anti money laundering, cryptocurrency, supervised classification, anomaly detection, active learning}


\settopmatter{printfolios=true}
\maketitle

\section{Introduction}
Money laundering is a high-impact problem on a global scale. Criminals obtain money illegally from serious crimes and inject it into the financial system as seemingly legitimate funds. Money laundering schemes usually involve large amounts of money and, when caught, typically result in large fines for financial institutions. Recent examples are the 1MDB ~\cite{1mdb_malaysia_2020} and the Danske Bank scandals~\cite{monroe_ousted_2020}. 

Governments and international organizations are building tighter regulations around money laundering and are broadening them to include cryptocurrencies~\cite{financial_crimes_enforcement_network_application_2019,european_union_directive_2018}, where criminals benefit from pseudonymity.

In the financial sector, Anti-Money Laundering (AML) efforts often rely on rule-based systems~\cite{Li2017}. However, vulnerabilities derive from the relative simplicity of publicly available rule-sets, leading to high false-positive rates (FPR) and low detection rates~\cite{Wang2008_intelligent}. Machine learning (ML) techniques overcome the rigidity of rule-based systems by inferring complex patterns from historical data, and can potentially increase detection rates and decrease FPRs. 

Recently, \citet{Weber2019} released a dataset, consisting of a sample of 200k labeled Bitcoin transactions, and evaluated various supervised models on it. Unfortunately, supervised methods are often unfeasible as institutions do not possess large-scale labeled datasets. This lack of labels is due to two main reasons. First, given the evolving complexity of money laundering schemes, it is unlikely to be possible to identify all (or even most) of the entities involved in money laundering. Second, labels resulting from law enforcement investigations are not immediate, and manual annotation is costly. Thus, in order to properly evaluate the practical feasibility of ML for AML, strategies that require no labels (unsupervised learning) or just a few labels (active learning) are paramount.

We address the real-world challenge of \emph{how to detect money laundering in a dataset with few labels}. Particularly, we show that:

\begin{enumerate}
    \item Detecting money laundering cases in the Bitcoin network without any labels is impossible since illicit transactions hide within clusters of licit behaviour (Section~\ref{sec:anomaly_detection_res}). 
    \item With just a few labels (approximately 5\% of the total), one can match the results of a supervised baseline by using Active Learning (AL)~(Section~\ref{sec:active_learning_res}). This setting mimics a real-world scenario with limited availability of human analysts for manual labeling.
\end{enumerate}

We extend the existing research on unsupervised illicit activity detection in cryptocurrency and financial transactions by benchmarking different methods on a real-world dataset with a relatively large number of positive cases. In this way, we overcome the typical limitation of evaluating on synthetic data or real data with few positive samples. Besides, to the best of our knowledge, we are the first work to apply AL to AML on a large transaction dataset and in the cryptocurrency setting, specifically.

We organize the remainder of the paper as follows. Section~\ref{sec:related_work} presents the related work. Section~\ref{sec:experimental_setup} details the experimental setup and introduces the relevant anomaly detection methods as well as AL concepts. In Section~\ref{sec:results} we present our results. Finally, the main conclusions are discussed in Section~\ref{sec:conclusions}.

\section{Related work}
\label{sec:related_work}
In this section, we present previous research on ML for AML in the context of both financial transactions and, more specifically, cryptocurrencies. For a thorough survey of ML approaches for AML, we refer the reader to ~\citet{Chen2018_review}.

Although approaches greatly vary, many methods assume money laundering cases to be outliers, i.e., illicit instances (a minority) should exhibit significantly different behaviours from legitimate ones (the majority). Typically, these approaches use unsupervised anomaly detection methods to model licit behaviour and find the instances that deviate from it~\cite{Tang2005,Liu2008,Wang2009,Zengan2009,Liu2010,Larik2011,Raza2011,Chen2014,Camino2017}.

Overall, the results of these studies are encouraging, reporting low FPRs~\cite{Liu2008,Liu2010,Chen2014} and good detection rates ~\cite{Liu2008,Chen2014,Zengan2009,Wang2009}. Some studies even report that the ML approaches were able to detect money laundering patterns that were previously unknown ~\cite{Tang2005} or not caught by rule-based systems ~\cite{Camino2017}. However, a fair comparison between methods is impossible, given the heterogeneity of the evaluation setups. In these studies, researchers use real-world datasets labeled by analysts~\cite{Liu2008,Liu2010,Camino2017}, with simulated illicit transactions~\cite{Tang2005,Zengan2009,Wang2009}, or no labels at all ~\cite{Larik2011, Raza2011}

Generally, authors are openly doubtful about real-world reproducibility of good results, in the face of intricate patterns and incomplete labels~\cite{Zengan2009,Wang2009,Chen2014}. The question arises on whether reliable anomaly detection is possible in non-synthetic data, as criminals could intentionally mimic normal behaviour. In our research, we contribute to assess the reproducibility of such results by conducting the first in-depth benchmark of anomaly detection methods in a labeled real-world cryptocurrency dataset and comparing their performance against a supervised baseline. 

Previous studies on money laundering in cryptocurrencies in particular are scarce and inconclusive due to a lack of labels for evaluation. Some conclude that supervised models perform well~\cite{Hu2019, Bartoletti2018, Monamo2016_multifaceted}. Others report low detection rates for unsupervised methods in extremely imbalanced data ~\cite{Wu2020,Pham2016_unsupervised,Pham2016_networkperspective,Monamo2016_multifaceted,Monamo2016}. Often, the evaluation of anomaly detection methods consists of checking whether the anomalies represent extreme cases~\cite{Pham2016_unsupervised,Pham2016_networkperspective} or behaviour deemed suspicious by human analysis~\cite{Hirshman2013}. 

Active Learning has been proposed as a method to reduce the number of labels needed for the training of an effective classifier by iteratively sampling the most informative samples for labeling from an initially unlabeled pool~\cite{Settles2009}. Given the apparent label scarcity in money laundering data, it is a highly relevant setting for the practical implementation of ML-based AML systems. Previously, \citet{Deng2009} applied AL to detect money laundering in financial transactions. In an account-level classification of 92 real-life accounts, they report that their method can accurately estimate the threshold hyperplane with only 22\% of the labels. AL has also successfully been applied in other fraud-related use-cases such as credit card fraud ~\cite{Carcillo1018} and network intrusion detection ~\cite{Almgren2004,Stokes2008,Gornitz2009}, reporting the sufficiency of as few as 1.5\% of the original labels to achieve near-optimal performance ~\cite{Gornitz2009}.

We conduct experiments with AL, assuming an unlabeled dataset and the capacity to acquire labels progressively to train a supervised classifier. We hereby extend the study by ~\citet{Deng2009} to a transaction-level analysis in a much larger cryptocurrency dataset.

\section{Experimental setup}
\label{sec:experimental_setup}

\subsection{Data}
\begin{figure}[!htb]
    \begin{center}
        \includegraphics[width=0.97\linewidth]{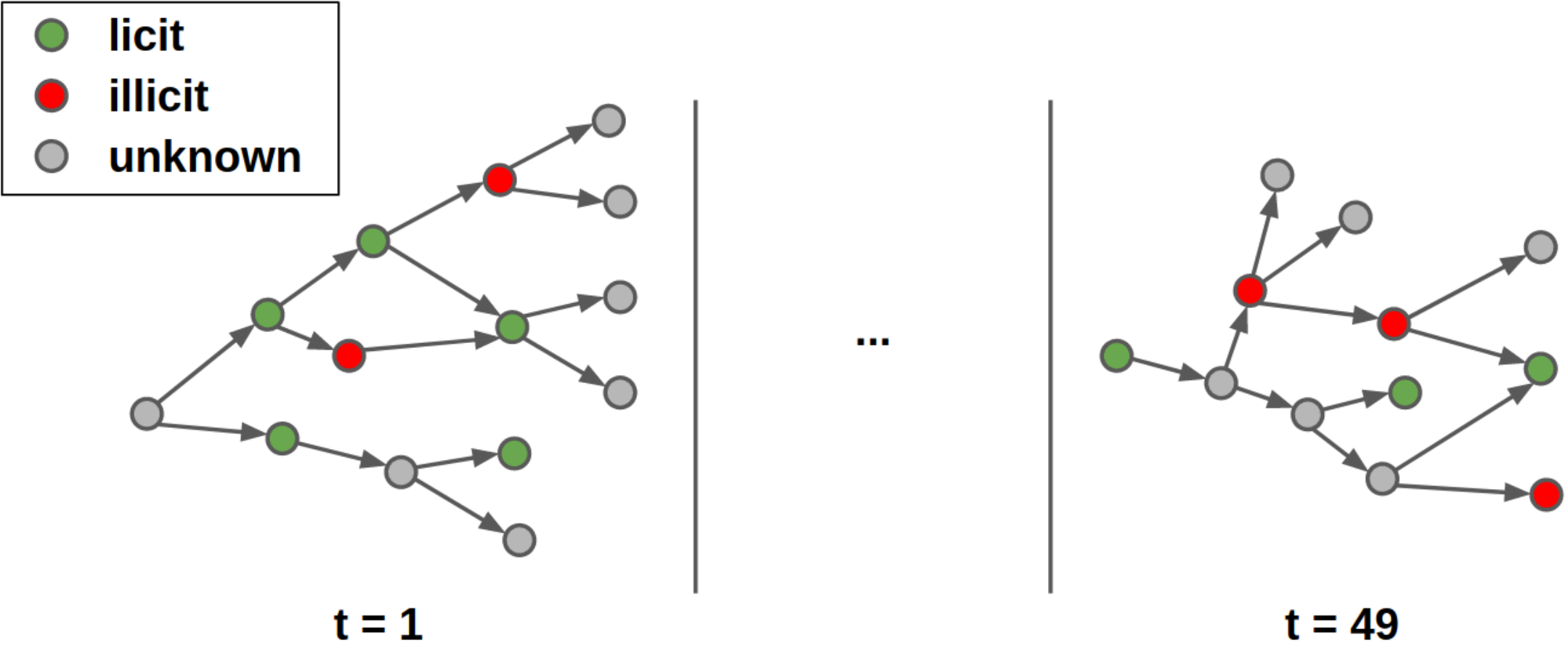}
    \end{center}
    \caption{Structure of the dataset (taken from~\citet{bellei_elliptic_2019}).}
    \label{fig:elliptic_dataset}
\end{figure}

We use the Bitcoin dataset\footnote{Available at https://www.kaggle.com/ellipticco/elliptic-data-set} released by Elliptic, a company dedicated to detecting financial crime in cryptocurrencies~\cite{Weber2019}. It includes 49 graphs sampled from the Bitcoin blockchain at different sequential moments in time (time-steps), as presented in Figure~\ref{fig:elliptic_dataset}. Each graph is a directed acyclic graph, starting from one transaction, and including subsequent related transactions on the blockchain, containing approximately two weeks of data.

Bitcoins transactions are transfers from one Bitcoin address (e.g., a person or company) to another, represented as nodes in the graph. Each transaction consumes the output of past transactions and generates outputs that can be spent by future transactions. The edges in the graph represent the flow of Bitcoins between transactions.

The dataset consists of 203,769 transactions, of which 21\% are labeled as licit, and 2\% as illicit, based on the \emph{category} of the bitcoin address that created the transaction. The remaining transactions are unlabeled. Illicit categories include scams, malware, terrorist organizations, and Ponzi schemes. Licit categories include exchanges, wallet providers, miners, and licit services. Each transaction has 166 features, 94 of which represent information about the transaction itself. The remaining features were constructed by~\citet{Weber2019} using information one-hop backward/forward from the transaction, such as the minimum, maximum, and standard deviation of each transaction feature. All features, except for the time-step, are fully anonymized and standardized with zero mean and unit variance.

\subsection{Methods}
In this section, we give an overview of the methods used in our experiments and discuss our experimental setup. Following ~\citet{Weber2019}, we split the data into sequential train and test datasets for all experiments. The train set includes all labeled samples up to the 34th time-step (29894 transactions), and the test set includes all labeled samples from the 35th time-step, inclusive, onward (16670 transactions). Like \citet{Weber2019} we evaluate all methods using the F1-score for the illicit class, hereafter referred to as illicit F1-score.

\subsubsection{Supervised Learning}
In order to benchmark unsupervised methods and AL, we first reproduce the results of~\citet{Weber2019} as our baseline.

We train each supervised model on the train set using all 166 features and then evaluate them on the entire test set. To measure performance over time, and following \citet{Weber2019}, we also report the illicit F1-score per time-step in the test set. We use the scikit-learn \cite{scikit-learn} implementation of logistic regression (LR) and random forest (RF) as well as the Python implementation of XGBoost~\cite{chen2016xgboost}. We present the results achieved using default parameters, as in \citet{Weber2019}. 

\subsubsection{Unsupervised Learning}
Anomaly detection methods are unsupervised learning techniques to detect outliers in a dataset. Literature suggests their effectiveness in the AML context (Section~\ref{sec:related_work}). For a thorough review of anomaly detection, we refer the reader to the surveys by~\citet{Chandola2009} and~\citet{Domingues2018}.

The standard definition of outliers refers to instances that are unlikely to be drawn from the same distribution as the train data or instances that are far from other data points in the feature space. Although we focus mainly on unsupervised anomaly detection, some methods are semi-supervised discriminators trained to learn a boundary around normal instances. In that context, outliers are instances that fall outside of the boundary~\cite{Zengan2009}. 

We test seven common anomaly detection algorithms with readily available Python implementations: Local Outlier Factor (LOF), K-Nearest Neighbours (KNN), Principal Component Analysis (PCA), One-Class Support Vector Machine (OCSVM), Cluster-based Outlier Factor (CBLOF), Angle-based Outlier Detection (ABOD), and Isolation Forest (IF). We aim at a diversity of strategies. We use the PyOD package implementations~\cite{zhao2019pyod} with default parameters. 

LOF and KNN start by computing the distance of each instance to its \emph{k} nearest-neighbour. Then, KNN defines that distance as the outlier score. LOF uses the distance to compute the instance's density, and if the density is substantially lower than the average density of its \emph{k} nearest-neighbours, the instance is declared anomalous. 

PCA and OCSVM define anomalies as observations that deviate from normal behaviour. They detect anomalous instances as observations with a large distance to the principal components (PCA) of non-anomalous observations or instances that lay outside of the decision boundary (OCSVM) learned around them. 

CBLOF uses the outcome of a clustering algorithm on the instances (in our case k-means) and classifies each cluster as either \emph{small} or \emph{large}. It calculates an anomaly score for each instance, marking instances that belong to small clusters or that are far from big clusters as anomalous. ABOD computes the pairwise cosine similarities between all points and classifies those with a low average radius and variance as anomalies. Lastly, IF isolates anomalies by performing recursive random splits on attribute values. Based on the resulting tree structure, anomalies are instances that are easy to isolate, i.e., have shorter paths.

The introduced methods use different anomaly scores and scales. Thus, a fair comparison requires evaluation at different \emph{contamination levels}, defined as the expected proportion of outliers in the dataset, and used to set a threshold for the decision function. Whereas the original PyOD implementation applies the contamination level on the scores of the train set, we apply it on the test set scores to guarantee that the desired percentage of positive cases (anomalies) in the test set is the same across methods. The contamination level here is analogous to a fixed alert rate in real AML systems, i.e., the percentage of cases flagged for further investigation by an analyst. We evaluate the illicit F1-score for each model at contamination levels between 0 and 1, with increments of 0.05. We also present the illicit F1-score of the RF supervised baseline, where we define the model threshold by setting the contamination level as the predicted positive rate (or alert rate), for comparison. 

\subsubsection{Active Learning}
AL is an incremental learning approach that interactively queries instances for labeling (e.g., by human analysts) and uses the increasing number of labeled instances to (re-)train a supervised model. It fits the AML context by addressing label scarcity and has previously been successfully applied to detect money laundering accounts based on financial transaction history. For an extensive survey on AL, we refer the reader to \citet{Settles2009}.

The goal of AL is to minimize the number of labels necessary to achieve adequate classifier performance. The process starts with a pool of unlabeled instances (the \emph{unlabeled pool}), although sometimes there is a residual number of labels. At each iteration, a \emph{query strategy} queries a batch of instances for manual labeling. After labeling, the instances go into the \emph{labeled pool}. Finally, a supervised algorithm (the \emph{classifier}) is trained on the labeled pool and evaluated on a test set. If the performance is not satisfactory, the querying process continues to enrich the labeled pool incrementally. To mimic the manual labeling process in our experiments, we append the labels to the queried instances.

In the literature, query strategies build on various models and uncertainty criteria. In this study, we focus on four query strategies trained on the labeled pool to find informative instances in the unlabeled pool. Two of them, uncertainty sampling and expected model change, are supervised, requiring an underlying supervised model to define queries. The other two, elliptic envelope and Isolation Forest (IF), are unsupervised and find outlying instances with regards to the labeled pool. We use random sampling as a baseline. This setup was based on previous work done on Feedzai's active learning annotation tool, which was used to run the experiments.

Expected model change~\cite{Settles2008, Settles2009} assumes that instances are more informative if they influence the model more strongly. It queries the unlabeled instances that lead to the most significant change in the model parameters by measuring the impact of labeling one unlabeled instance on the gradient of the model's loss function. Thus, this strategy applies only to gradient-based classifiers. In our experiment, we use LR. The expected model change is a weighted sum over all possible labels since the labels of the instances are unknown before querying. Then, at each iteration, we query the labels of the instances with the largest expected gradients. 

Uncertainty sampling is one of the most commonly used query strategies~\cite{Lewis1994, Settles2009}. It queries the instances about which a model is most uncertain. Assuming a probabilistic learning model and a binary classification problem, this translates to querying the instances with the predicted score closest to 0.5. In our study, we use the same type of classifier for uncertainty sampling and evaluating on the test set; for instance, if the classifier is RF, we also conduct uncertainty sampling using RF.

The two unsupervised query strategies used are IF and elliptic envelope. Outliers are transactions with high anomaly scores (IF) or a large Mahalanobis distance to a multivariate Gaussian distribution fit on the labeled pool (elliptic envelope).

We combine unsupervised and supervised query strategies in our experiments, depending on the number of illicit instances in the labeled pool. After an initial random sample of one batch of instances, we use an unsupervised \emph{warm-up learner} that samples instances until the labeled pool includes at least one illicit instance. When we reach this threshold, we either switch to a supervised \emph{hot learner} or continue to use the warm-up learner. As the classifier, we use the three supervised models evaluated in the supervised baselines: RF, XGBoost, and LR. We compare all AL setups against a baseline that queries random instances at each iteration (also used as a warm-up learner). We use a batch size of 50 instances sampled at each iteration for all experiments. Each AL setup is run five times with different random seeds to ensure the robustness of the results. We assess the performance of each AL setup through the median illicit F1-score and the confidence intervals at each labeled pool size. This setup and parameter choices follow the work by~\citet{AL_Ascensao_et_al} developed at Feedzai concurrently.

\section{Results}
\label{sec:results}
In this section, we present the experimental results for the supervised baseline, followed by the anomaly detection and the AL benchmarks.

\subsection{Supervised baselines}
\label{sec:supervised_baseline_res}
We are able to reproduce the results presented by~\citet{Weber2019} closely. Over five runs (with different seeds), we achieve an illicit F1-score on the test set of 0.76 for XGBoost, 0.45 for LR, and 0.83 for RF. Thus, the best supervised baseline is achieved with the RF model. As \citet{Weber2019}, we observe that model performance is profoundly affected by a sudden dark market shutdown at time-step~43.

\begin{figure}[!t]
    \begin{center}
        \includegraphics[width=0.97\linewidth]{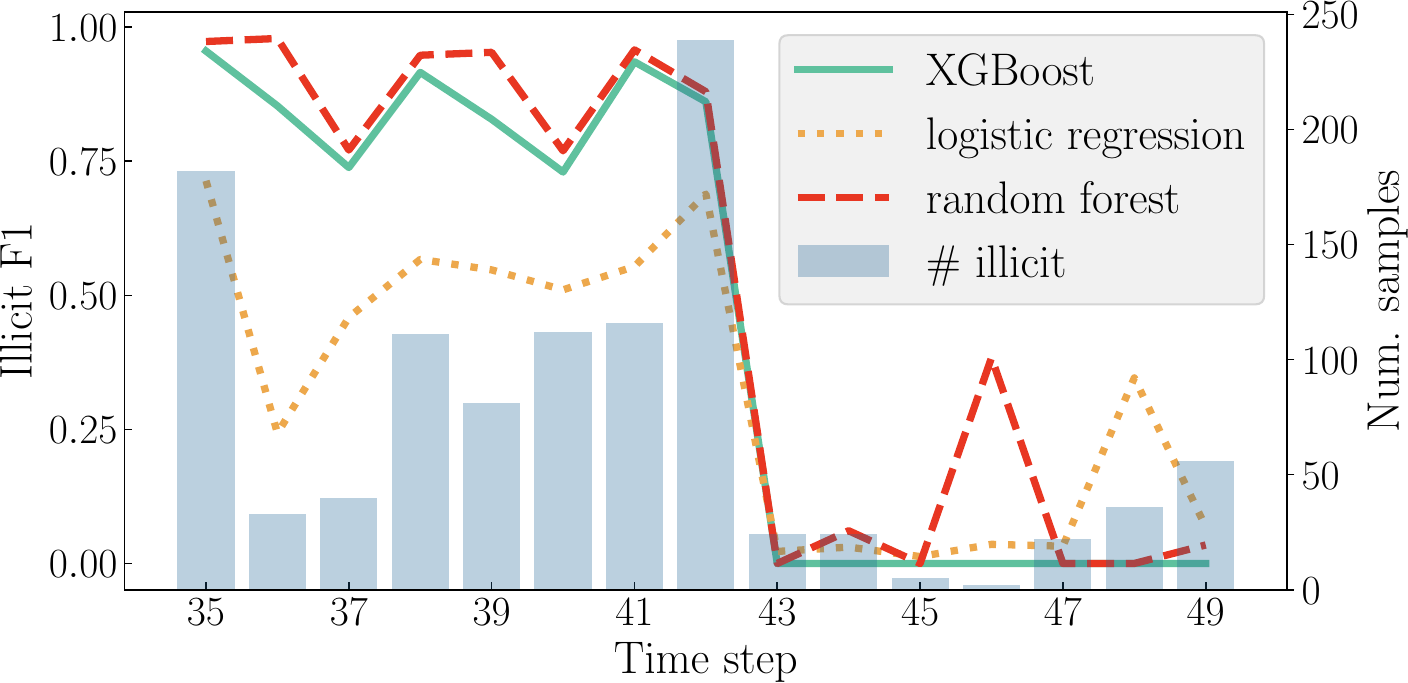}
    \end{center}
    \caption{Illicit F1-score for each supervised baseline, across time-steps.}
    \label{fig:supervised_baseline}
\end{figure}

\subsection{Anomaly detection}
\label{sec:anomaly_detection_res}
In Table~\ref{tab:f1_per_contamination}, we present the illicit F1-score of the explored anomaly detection methods as well as the RF supervised baseline at different contamination levels. Recall that, at each contamination level, we define the threshold of the RF model so that it leads to an alert rate equal to that contamination level.

\begin{table}[!b]
    \centering
    \caption{Anomaly detection methods illicit F1-score by contamination level (RF supervised baseline for reference).}
    \label{tab:f1_per_contamination}
    \begin{tabular}{|l|llll|}
        \hline
        \multirow{2}{*}{\textbf{Model}} & \multicolumn{4}{c|}{\textbf{Contamination level}} \\
         & \textbf{0.05} & \textbf{0.1} & \textbf{0.15} & \textbf{0.2}  \\ 
    \hline
        RF supervised baseline & 0.82 & 0.58 & 0.46 & 0.39 \\ \hline
        LOF & 0.11 & 0.15 & 0.19 & 0.18  \\ \hline
        ABOD & 0.07 & 0.07 & 0.07 & 0.07  \\ \hline
        KNN & 0.03 & 0.04 & 0.05 & 0.06  \\ \hline
        OCSVM & 0.01 & 0.03 & 0.03 & 0.04 \\ \hline
        CBLOF & 0.01 & 0.02 & 0.03 & 0.04  \\ \hline
        PCA & 0.01 & 0.01 & 0.02 & 0.02  \\ \hline
        IF & 0.00 & 0.00 & 0.00 & 0.01  \\ \hline
    \end{tabular}
\end{table}

\begin{figure}[!t]
    \begin{center}
        \includegraphics[width=0.97\linewidth]{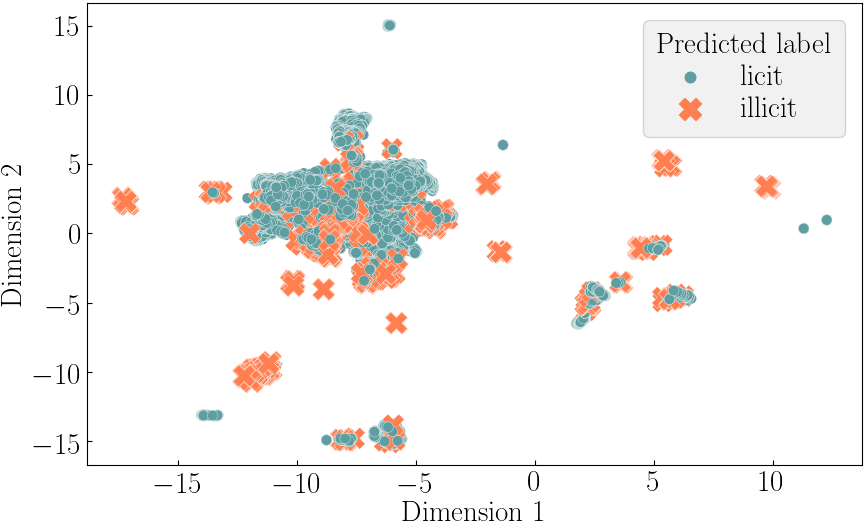}
    \end{center}
    \caption{UMAP projection of the test set, colored by the labels predicted by IF.}
    \label{fig:umap_predicted}
\end{figure}

\begin{figure}[!t]
    \begin{center}
        \includegraphics[width=0.97\linewidth]{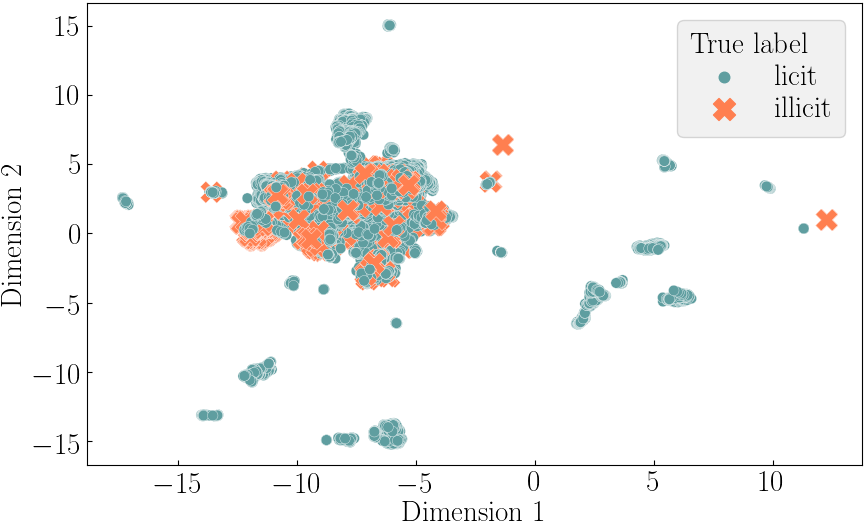}
    \end{center}
    \caption{UMAP projection of the test set, colored by the true labels.}
    \label{fig:umap_true}
\end{figure}

Anomaly detection methods perform significantly below the RF supervised baseline across all contamination levels. These results are not consistent with past studies, where anomaly detection methods perform adequately for AML (Section~\ref{sec:related_work}). However, we note that these studies often use synthetically generated anomalous data points that are outliers by design. Furthermore, there could be differences between money laundering patterns in financial transactions and Bitcoin transfers. In the real-life Bitcoin transaction dataset, we see that illicit cases are indeed not outlying.

To illustrate this, we apply the Uniform Manifold Approximation and Projection (UMAP)~\cite{McInnes2018} to the test set and build two plots with the resulting projection. In the first (Figure \ref{fig:umap_predicted}), we color each observation based on the predicted label of the worst-performing method (IF), while in the second (Figure \ref{fig:umap_true}), we color each observation based on the true label. We can then see that the IF classifies most outlying instances as illicit, as intended. Still, the true labels presented in Figure \ref{fig:umap_true} reveal that the illicit instances in the dataset are not actually outlying, but instead hiding among licit transactions.

The observation that not all outliers are illicit and that not all illicit transactions are outliers is reasonable in AML as sophisticated criminals obfuscate their activity by mimicking normal behaviour, hiding in regions of high nominal density. This problem was previously acknowledged by \citet{Das2016}. Thus, we conclude that anomaly detection methods are ineffective for the unsupervised classification task in this real-life Bitcoin dataset. 

\subsection{Active learning}
\label{sec:active_learning_res}
Table \ref{tab:active_learning} summarises the results of the AL benchmark for each of the three different classifiers used for the supervised baselines. We conclude that switching to a supervised hot-learner significantly improves performance over the continued use of an unsupervised warm-up learners. Among hot-learners, however, there is no clear best policy.

Furthermore, we can see that random sampling as the warm-up learner leads to a faster improvement in model performance (i.e., better performance for smaller labeled pool sizes) compared to anomaly detection methods. This observation aligns with previous considerations on the poor performance of anomaly detection methods (Section~\ref{sec:anomaly_detection_res}). Since these methods fail to detect illicit instances, they are ineffective at querying illicit instances to be added to the labeled pool to improve the performance of a supervised classifier quickly. We observe that elliptic envelope performs above IF (also consistent with previous results).

Table \ref{tab:active_learning} additionally shows that XGBoost and LR temporarily surpass their supervised baseline, i.e., the performance they achieved when trained on the entire train set (Section \ref{sec:supervised_baseline_res}. The classifiers perform better when trained only on a sample of the labeled data but eventually converge to their supervised baseline as the labeled pool increases over time. This result can be because the labeled pool consists of the most relevant samples at the beginning of the AL process and, at the same time, the class imbalance increases over time. \citet{Laws2008} acknowledge that, in some cases, early stopping of an AL process might prevent this model degradation. Note, however, that even if XGBoost and LR surpass their own supervised baselines, they do not surpass the best supervised baseline, which was achieved with the RF model.

\begin{table*}[!htb]
\centering
\caption{Average illicit F1-score over five runs for each AL setup, consisting of an unsupervised warm-up learner, an optional supervised hot learner and the classifier that is evaluated on the test set. We compare the results to each classifier's respective supervised baseline (Section \ref{sec:supervised_baseline_res}). Results are ordered by the illicit F1-score with 3000 labels. Best values for each labeled pool size across classifiers are highlighted in bold.}
\begin{tabular}{|l|l|l|lllll|l|}
\hline
\multicolumn{2}{|c|}{\textbf{Query strategies}}  & \multirow{3}{*}{\textbf{Classifier}} & \multicolumn{5}{c|}{\textbf{Labeled pool size}} & \multirow{3}{*}{\textbf{Supervised baseline}}  \\
\multirow{2}{*}{\textbf{Warm-up learner}}   & \multirow{2}{*}{\textbf{Hot learner}}&  & \textbf{200} & \textbf{500} &\textbf{1000}&\textbf{1500}&\textbf{3000}&\\ 
& & &  \textbf{(0.7\%)} & \textbf{(1.7\%)} &\textbf{(3.3\%)}&\textbf{(5\%)}&\textbf{(10\%)}&\\ \hline
isolation forest   & uncertainty sampling  & \multirow{9}{*}{random forest}  & 0.75   & 0.75   & 0.80& \textbf{0.82}  & \textbf{0.83}  &\\
random sampling& uncertainty sampling  &  & 0.73   & 0.75   & \textbf{0.81}  & \textbf{0.82}  & 0.82&\\
elliptic envelope  & uncertainty sampling  &  & 0.65   & \textbf{0.77} & 0.80& \textbf{0.82}  & 0.82&\\
isolation forest   & expected model change && 0.56   & 0.61   & 0.77& 0.79& 0.81&    \\
random sampling& expected model change &  & \textbf{0.76} & \textbf{0.77} & 0.78& 0.78& 0.81& 0.83 \\
elliptic envelope  & expected model change &  & 0.60   & 0.72   & 0.76& 0.77& 0.81&\\
random sampling& --&  & 0.74   & 0.76   & 0.76& 0.78& 0.80&\\
elliptic envelope  & --&  & 0.50   & 0.53   & 0.56& 0.65& 0.70&\\
isolation forest   & --&  & 0.67   & 0.65   & 0.59& 0.63& 0.62&\\ \hline
isolation forest   & uncertainty sampling  & \multirow{9}{*}{XGBoost} & 0.67   & \textbf{0.77} & 0.80& 0.79& 0.80&\\
elliptic envelope  & expected model change &  & 0.65   & 0.75   & 0.77& 0.75& 0.79&\\
random sampling& expected model change &  & 0.70   & 0.75   & 0.79& 0.80& 0.78&\\
isolation forest   & expected model change &  & 0.60   & 0.75   & 0.77& 0.76& 0.75&   \\
elliptic envelope  & --&  & 0.53   & 0.64   & 0.53& 0.61& 0.68& 0.76  \\
elliptic envelope  & uncertainty sampling  &  & 0.62   & 0.62   & 0.64& 0.80& 0.64&\\
random sampling& uncertainty sampling  &  & 0.72   & 0.76   & 0.64& 0.60& 0.64&\\
random sampling& --&  & 0.66   & 0.58   & 0.75& 0.74& 0.59&\\
isolation forest   & --&  & 0.38   & 0.38   & 0.46& 0.44& 0.57&\\ \hline
isolation forest  & expected model change & \multirow{9}{*}{logistic regression} & 0.22   & 0.59   & 0.63& 0.66& 0.62&\\
elliptic envelope& expected model change &  & 0.20   & 0.48   & 0.61& 0.61& 0.61&\\
random sampling  & expected model change &  & 0.44   & 0.54   & 0.58& 0.64& 0.60&\\
elliptic envelope & uncertainty sampling  &  & 0.41   & 0.52   & 0.63& 0.63& 0.60&  \\
isolation forest & uncertainty sampling  &  & 0.37   & 0.53   & 0.61& 0.60& 0.58& 0.45 \\
random sampling  & uncertainty sampling  &  & 0.40   & 0.50   & 0.57& 0.58& 0.55&\\
random sampling  & --&  & 0.36   & 0.36   & 0.36& 0.37& 0.39&\\
elliptic envelope & --&  & 0.28   & 0.25   & 0.24& 0.24& 0.22&\\
isolation forest  & --&  & 0.25   & 0.24   & 0.29& 0.21& 0.02&\\
 \hline
\end{tabular}
\label{tab:active_learning}
\end{table*}

\begin{figure}[!t]
    \begin{center}
        \includegraphics[width=0.97\linewidth]{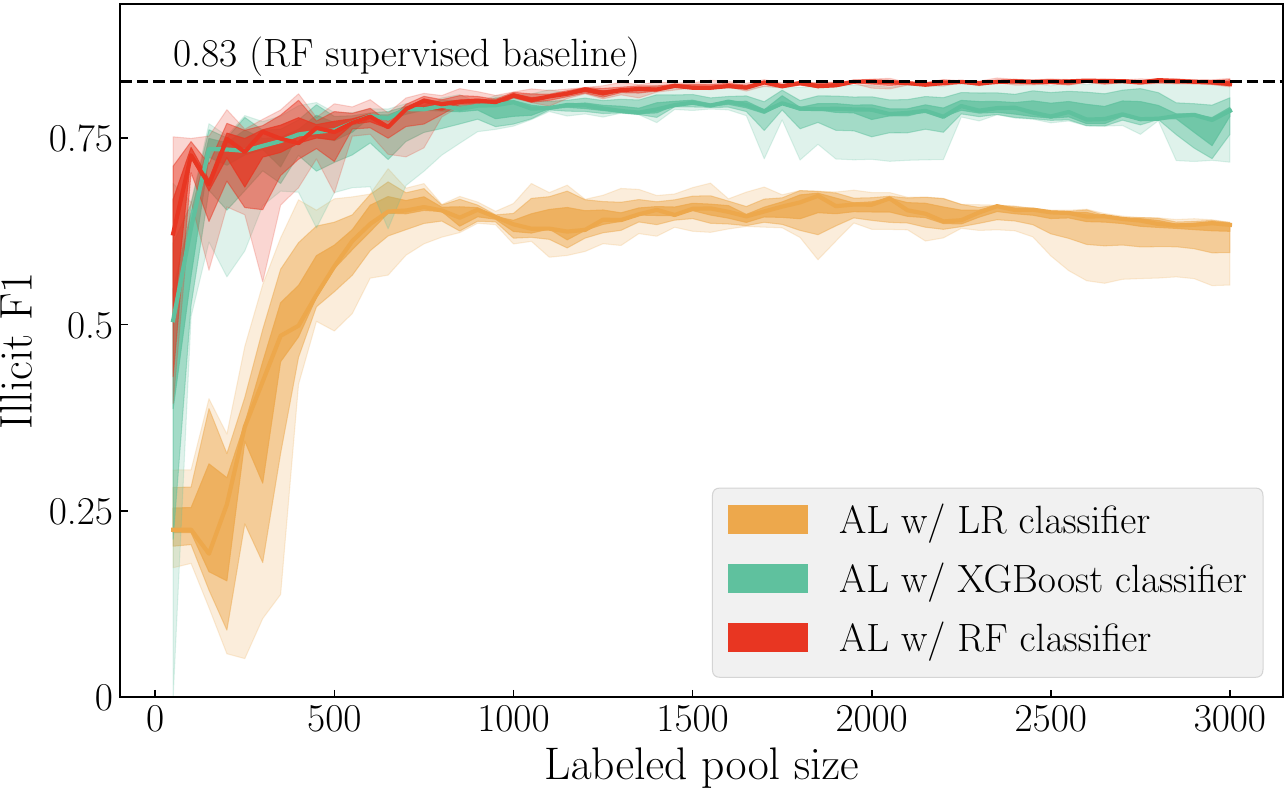}
    \end{center}
    \caption{Best AL setups for each classifier and the RF supervised baseline.}
    \label{fig:AL_ds_default}
\end{figure}

\begin{figure}[!t]
    \begin{center}
        \includegraphics[width=0.97\linewidth]{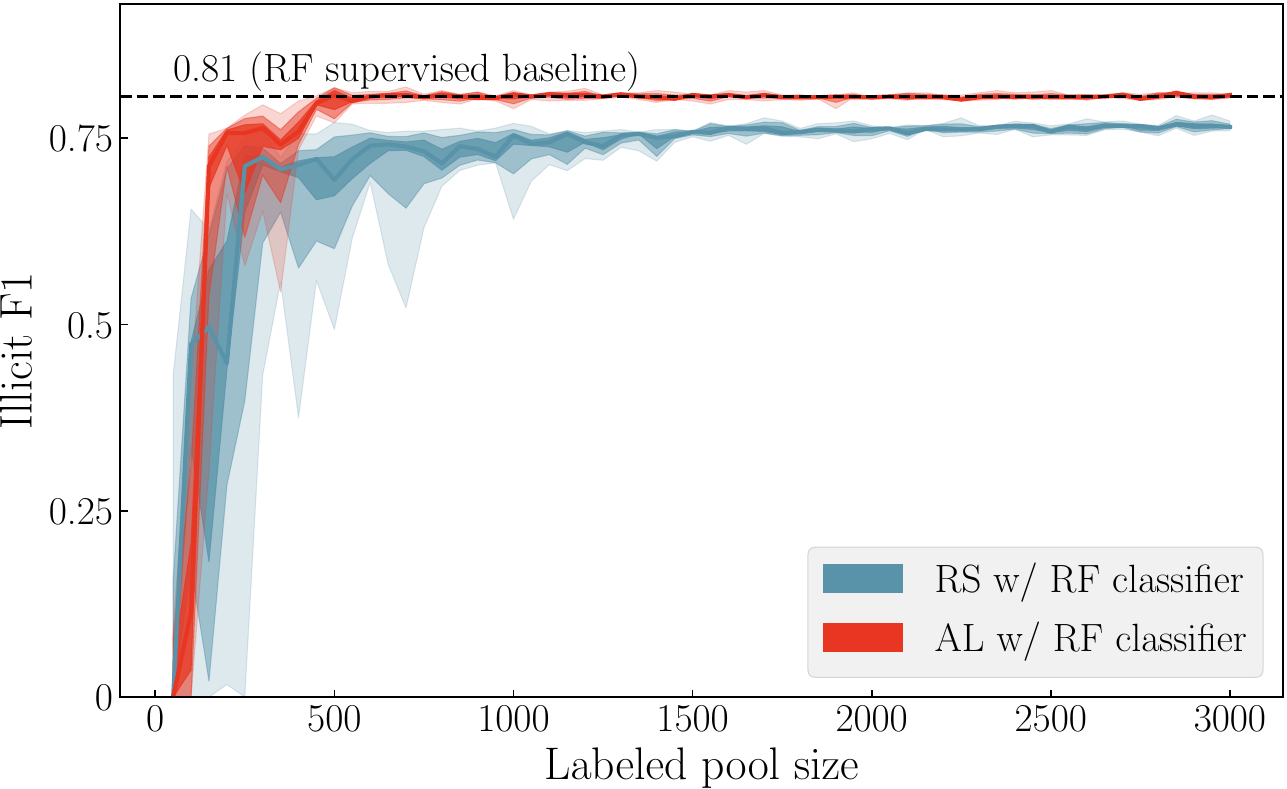}
    \end{center}
    \caption{AL versus Random Sampling (RS) with a RF classifier and the RF supervised baseline at 2\% illicit rate.}
    \label{fig:AL_ds_2_perc}
\end{figure}

\begin{figure}[!t]
    \begin{center}
        \includegraphics[width=0.97\linewidth]{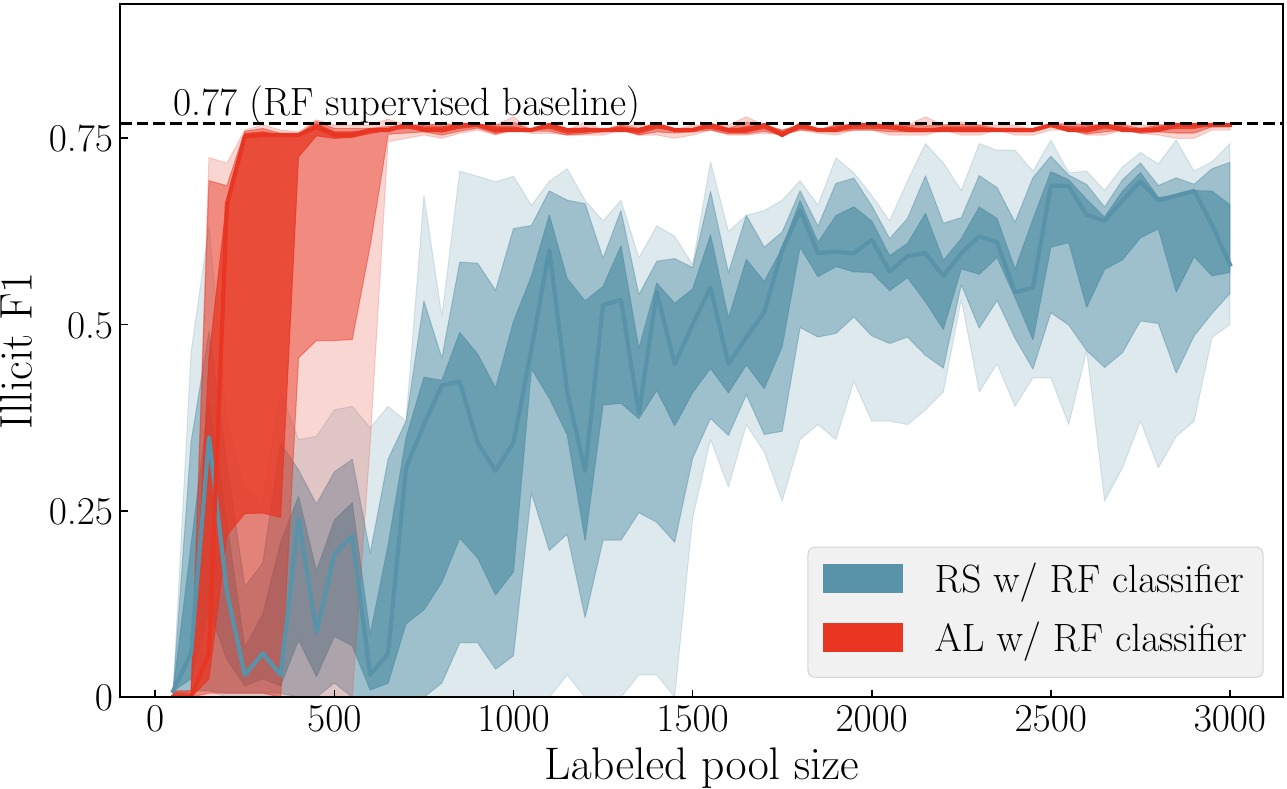}
    \end{center}
    \caption{AL versus Random Sampling (RS) with a RF classifier and the RF supervised baseline at 0.5\% illicit rate.}
    \label{fig:AL_ds_05_perc}
\end{figure}

Figure \ref{fig:AL_ds_default} shows the performance over time of the best AL setup for the three classifiers tested. For comparison, it also includes the performance achieved by the best supervised baseline, the RF supervised baseline. With the presented AL setup, all classifiers stabilize after 1000 labels, with RF and XGBoost exhibiting faster performance increase. RF reaches its baseline's performance with only 5\% of the original labels, or 1500 out of the original 30000 labels (Figure~\ref{fig:AL_ds_default}). We can even see a near-optimal performance with as few as 500 labels.

From Table \ref{tab:active_learning}, we can observe that the random sampling baseline achieves a similar performance to the more sophisticated AL strategies. Our intuition is that the classifier will start approaching good performances when the labeled pool includes a sufficient number of illicit instances and, because the dataset has approximately 10\% of illicit cases, random sampling can quickly reach that sufficient number.

In reality, financial crime is extremely rare among licit transactions, and thus datasets are highly imbalanced ~\citet{sudjianto2010statistical}. Since we are interested in the practical relevance of AL, we compare the best performing AL setup against random sampling in a dataset with a higher, more realistic class imbalance. Specifically, we apply a random undersampling of the minority class of the Ellipic dataset to achieve illicit rates of 2\% and 0.5\%. The results are plotted in Figures~\ref{fig:AL_ds_2_perc} and ~\ref{fig:AL_ds_05_perc}, respectively. For comparison, we indicate the RF supervised baseline performance at the respective reduced fraud rates.

As expected, the AL query strategies increasingly outperform random sampling as imbalance increases. For highly imbalanced datasets, the best setup uses random sampling (warm-up) followed by uncertainty sampling (hot learner).

\section{Conclusion}
\label{sec:conclusions}
In this study, we conducted experiments to detect illicit activity on the Bitcoin transaction dataset released by Elliptic. Using a supervised setting similar to \citet{Weber2019} as our baseline, we studied the detection ability of machine learning models in a more realistic setting with restricted access to labels, using unsupervised methods, and Active Learning (AL).

Our results indicate that unsupervised anomaly detection methods have poor performance, and we present evidence that anomalies in the feature-space are not indicative of illicit behaviour. This finding highlights that experiments conducted on (partially) synthetic data can be misleading and emphasizes the importance of conducting experiments on real-life datasets to draw reliable conclusions.

To improve upon the unsupervised performance, we studied the case where few labels can be obtained by using AL and determined the minimum amount of labeled instances necessary to achieve a performance close to the best supervised baseline. This setting is realistic and akin to asking money laundering analysts to review cases that an AL model indicates as informative. We obtained similar performance to the best supervised baseline by using just a few hundred labels (5\% of the total).

It remains to explore if the distribution of classes that we found in the Bitcoin dataset holds for other real-life datasets and different labeling strategies. Furthermore, given the need for proper AML processes in the entire financial system, it is crucial to conduct similar benchmarks on other verticals such as bank transfers, deposits or loans, using real datasets with proper labels.

\section*{Acknowledgments}

We want to thank the entire Feedzai Research team, who always gave insightful suggestions. In particular, we want to give special thanks to Marco Sampaio, Ricardo Barata, and Miguel Leite for their support with the AL experiments. 

\subsection*{Note on reproducibility}
The code to reproduce the results presented in Sections \ref{sec:supervised_baseline_res} and \ref{sec:anomaly_detection_res} is available online at \url{https://github.com/feedzai/research-aml-elliptic}. We do not include the code for the AL experiments for intellectual property purposes.

\bibliographystyle{ACM-Reference-Format}
\bibliography{reaml}


\begin{thebibliography}{44}


\ifx \showCODEN    \undefined \def \showCODEN     #1{\unskip}     \fi
\ifx \showDOI      \undefined \def \showDOI       #1{#1}\fi
\ifx \showISBNx    \undefined \def \showISBNx     #1{\unskip}     \fi
\ifx \showISBNxiii \undefined \def \showISBNxiii  #1{\unskip}     \fi
\ifx \showISSN     \undefined \def \showISSN      #1{\unskip}     \fi
\ifx \showLCCN     \undefined \def \showLCCN      #1{\unskip}     \fi
\ifx \shownote     \undefined \def \shownote      #1{#1}          \fi
\ifx \showarticletitle \undefined \def \showarticletitle #1{#1}   \fi
\ifx \showURL      \undefined \def \showURL       {\relax}        \fi
\providecommand\bibfield[2]{#2}
\providecommand\bibinfo[2]{#2}
\providecommand\natexlab[1]{#1}
\providecommand\showeprint[2][]{arXiv:#2}

\bibitem[\protect\citeauthoryear{Almgren and Jonsson}{Almgren and
  Jonsson}{2004}]%
        {Almgren2004}
\bibfield{author}{\bibinfo{person}{Magnus Almgren} {and}
  \bibinfo{person}{Erland Jonsson}.} \bibinfo{year}{2004}\natexlab{}.
\newblock \showarticletitle{{Using active learning in intrusion detection}}.
\newblock \bibinfo{journal}{\emph{Proceedings of the Computer Security
  Foundations Workshop}}  \bibinfo{volume}{17}, \bibinfo{pages}{88--98}.
\newblock


\bibitem[\protect\citeauthoryear{Barata, Leite, Pacheco, Sampaio, Ascensão,
  and Bizarro}{Barata et~al\mbox{.}}{2021}]%
        {AL_Ascensao_et_al}
\bibfield{author}{\bibinfo{person}{Ricardo Barata}, \bibinfo{person}{Miguel
  Leite}, \bibinfo{person}{Ricardo Pacheco}, \bibinfo{person}{Marco O.~P.
  Sampaio}, \bibinfo{person}{João~Tiago Ascensão}, {and}
  \bibinfo{person}{Pedro Bizarro}.} \bibinfo{year}{2021}\natexlab{}.
\newblock \bibinfo{title}{Active learning for online training in imbalanced
  data streams under cold start}.
\newblock
\newblock
\showeprint[arxiv]{cs.LG/2107.07724}


\bibitem[\protect\citeauthoryear{Bartoletti, Pes, and Serusi}{Bartoletti
  et~al\mbox{.}}{2018}]%
        {Bartoletti2018}
\bibfield{author}{\bibinfo{person}{Massimo Bartoletti},
  \bibinfo{person}{Barbara Pes}, {and} \bibinfo{person}{Sergio Serusi}.}
  \bibinfo{year}{2018}\natexlab{}.
\newblock \showarticletitle{Data mining for detecting Bitcoin Ponzi schemes}.
  In \bibinfo{booktitle}{\emph{2018 Crypto Valley Conference on Blockchain
  Technology (CVCBT)}}. IEEE, \bibinfo{pages}{75--84}.
\newblock


\bibitem[\protect\citeauthoryear{Bellei}{Bellei}{2019}]%
        {bellei_elliptic_2019}
\bibfield{author}{\bibinfo{person}{Claudio Bellei}.}
  \bibinfo{year}{2019}\natexlab{}.
\newblock \showarticletitle{The Elliptic Data Set: opening up machine learning
  on the blockchain}.
\newblock \bibinfo{journal}{\emph{Medium}} (\bibinfo{date}{Aug.}
  \bibinfo{year}{2019}).
\newblock
\urldef\tempurl%
\url{https://medium.com/elliptic/the-elliptic-data-set-opening-up-machine-learning-on-the-blockchain-e0a343d99a14}
\showURL{%
\tempurl}


\bibitem[\protect\citeauthoryear{Camino, State, Montero, and Valtchev}{Camino
  et~al\mbox{.}}{2017}]%
        {Camino2017}
\bibfield{author}{\bibinfo{person}{Ramiro~Daniel Camino}, \bibinfo{person}{Radu
  State}, \bibinfo{person}{Leandro Montero}, {and} \bibinfo{person}{Petko
  Valtchev}.} \bibinfo{year}{2017}\natexlab{}.
\newblock \showarticletitle{Finding Suspicious Activities in Financial
  Transactions and Distributed Ledgers}. In \bibinfo{booktitle}{\emph{2017 IEEE
  International Conference on Data Mining Workshops (ICDMW)}}. IEEE,
  \bibinfo{pages}{787--796}.
\newblock


\bibitem[\protect\citeauthoryear{Carcillo, Le~Borgne, Caelen, and
  Bontempi}{Carcillo et~al\mbox{.}}{2018}]%
        {Carcillo1018}
\bibfield{author}{\bibinfo{person}{Fabrizio Carcillo},
  \bibinfo{person}{Yann-A{\"e}l Le~Borgne}, \bibinfo{person}{Olivier Caelen},
  {and} \bibinfo{person}{Gianluca Bontempi}.} \bibinfo{year}{2018}\natexlab{}.
\newblock \showarticletitle{Streaming active learning strategies for real-life
  credit card fraud detection: assessment and visualization}.
\newblock \bibinfo{journal}{\emph{International Journal of Data Science and
  Analytics}} \bibinfo{volume}{5}, \bibinfo{number}{4} (\bibinfo{year}{2018}),
  \bibinfo{pages}{285--300}.
\newblock


\bibitem[\protect\citeauthoryear{Chandola, Banerjee, and Kumar}{Chandola
  et~al\mbox{.}}{2009}]%
        {Chandola2009}
\bibfield{author}{\bibinfo{person}{Varun Chandola}, \bibinfo{person}{Arindam
  Banerjee}, {and} \bibinfo{person}{Vipin Kumar}.}
  \bibinfo{year}{2009}\natexlab{}.
\newblock \showarticletitle{Anomaly detection: A survey}.
\newblock \bibinfo{journal}{\emph{ACM computing surveys (CSUR)}}
  \bibinfo{volume}{41}, \bibinfo{number}{3} (\bibinfo{year}{2009}),
  \bibinfo{pages}{1--58}.
\newblock


\bibitem[\protect\citeauthoryear{Chen and Guestrin}{Chen and Guestrin}{2016}]%
        {chen2016xgboost}
\bibfield{author}{\bibinfo{person}{Tianqi Chen} {and} \bibinfo{person}{Carlos
  Guestrin}.} \bibinfo{year}{2016}\natexlab{}.
\newblock \showarticletitle{Xgboost: A scalable tree boosting system}. In
  \bibinfo{booktitle}{\emph{Proceedings of the 22nd acm sigkdd international
  conference on knowledge discovery and data mining}}.
  \bibinfo{pages}{785--794}.
\newblock


\bibitem[\protect\citeauthoryear{Chen, Nazir, Teoh, Karupiah,
  et~al\mbox{.}}{Chen et~al\mbox{.}}{2014}]%
        {Chen2014}
\bibfield{author}{\bibinfo{person}{Zhiyuan Chen}, \bibinfo{person}{Amril
  Nazir}, \bibinfo{person}{Ee~Na Teoh}, \bibinfo{person}{Ettikan~Kandasamy
  Karupiah}, {et~al\mbox{.}}} \bibinfo{year}{2014}\natexlab{}.
\newblock \showarticletitle{Exploration of the effectiveness of expectation
  maximization algorithm for suspicious transaction detection in anti-money
  laundering}. In \bibinfo{booktitle}{\emph{2014 IEEE Conference on Open
  Systems (ICOS)}}. IEEE, \bibinfo{pages}{145--149}.
\newblock


\bibitem[\protect\citeauthoryear{Chen, {Van Khoa}, Teoh, Nazir, Karuppiah, and
  Lam}{Chen et~al\mbox{.}}{2018}]%
        {Chen2018_review}
\bibfield{author}{\bibinfo{person}{Zhiyuan Chen}, \bibinfo{person}{Le~Dinh {Van
  Khoa}}, \bibinfo{person}{Ee~Na Teoh}, \bibinfo{person}{Amril Nazir},
  \bibinfo{person}{Ettikan~Kandasamy Karuppiah}, {and} \bibinfo{person}{Kim~Sim
  Lam}.} \bibinfo{year}{2018}\natexlab{}.
\newblock \showarticletitle{{Machine learning techniques for anti-money
  laundering (AML) solutions in suspicious transaction detection: a review}}.
\newblock \bibinfo{journal}{\emph{Knowledge and Information Systems}}
  \bibinfo{volume}{57}, \bibinfo{number}{2} (\bibinfo{year}{2018}),
  \bibinfo{pages}{245--285}.
\newblock


\bibitem[\protect\citeauthoryear{Das, Wong, Dietterich, Fern, and Emmott}{Das
  et~al\mbox{.}}{2016}]%
        {Das2016}
\bibfield{author}{\bibinfo{person}{Shubhomoy Das}, \bibinfo{person}{Weng-Keen
  Wong}, \bibinfo{person}{Thomas Dietterich}, \bibinfo{person}{Alan Fern},
  {and} \bibinfo{person}{Andrew Emmott}.} \bibinfo{year}{2016}\natexlab{}.
\newblock \showarticletitle{Incorporating expert feedback into active anomaly
  discovery}. In \bibinfo{booktitle}{\emph{2016 IEEE 16th International
  Conference on Data Mining (ICDM)}}. IEEE, \bibinfo{pages}{853--858}.
\newblock


\bibitem[\protect\citeauthoryear{Deng, Joseph, Sudjianto, and Wu}{Deng
  et~al\mbox{.}}{2009}]%
        {Deng2009}
\bibfield{author}{\bibinfo{person}{Xinwei Deng}, \bibinfo{person}{V~Roshan
  Joseph}, \bibinfo{person}{Agus Sudjianto}, {and} \bibinfo{person}{CF~Jeff
  Wu}.} \bibinfo{year}{2009}\natexlab{}.
\newblock \showarticletitle{Active learning through sequential design, with
  applications to detection of money laundering}.
\newblock \bibinfo{journal}{\emph{J. Amer. Statist. Assoc.}}
  \bibinfo{volume}{104}, \bibinfo{number}{487} (\bibinfo{year}{2009}),
  \bibinfo{pages}{969--981}.
\newblock


\bibitem[\protect\citeauthoryear{Domingues, Filippone, Michiardi, and
  Zouaoui}{Domingues et~al\mbox{.}}{2018}]%
        {Domingues2018}
\bibfield{author}{\bibinfo{person}{R{\'e}mi Domingues},
  \bibinfo{person}{Maurizio Filippone}, \bibinfo{person}{Pietro Michiardi},
  {and} \bibinfo{person}{Jihane Zouaoui}.} \bibinfo{year}{2018}\natexlab{}.
\newblock \showarticletitle{A comparative evaluation of outlier detection
  algorithms: Experiments and analyses}.
\newblock \bibinfo{journal}{\emph{Pattern Recognition}}  \bibinfo{volume}{74}
  (\bibinfo{year}{2018}), \bibinfo{pages}{406--421}.
\newblock


\bibitem[\protect\citeauthoryear{Gao}{Gao}{2009}]%
        {Zengan2009}
\bibfield{author}{\bibinfo{person}{Zengan Gao}.}
  \bibinfo{year}{2009}\natexlab{}.
\newblock \showarticletitle{Application of cluster-based local outlier factor
  algorithm in anti-money laundering}. In \bibinfo{booktitle}{\emph{2009
  International Conference on Management and Service Science}}. IEEE,
  \bibinfo{pages}{1--4}.
\newblock


\bibitem[\protect\citeauthoryear{G{\"o}rnitz, Kloft, Rieck, and
  Brefeld}{G{\"o}rnitz et~al\mbox{.}}{2009}]%
        {Gornitz2009}
\bibfield{author}{\bibinfo{person}{Nico G{\"o}rnitz}, \bibinfo{person}{Marius
  Kloft}, \bibinfo{person}{Konrad Rieck}, {and} \bibinfo{person}{Ulf Brefeld}.}
  \bibinfo{year}{2009}\natexlab{}.
\newblock \showarticletitle{Active learning for network intrusion detection}.
  In \bibinfo{booktitle}{\emph{Proceedings of the 2nd ACM workshop on Security
  and artificial intelligence}}. \bibinfo{pages}{47--54}.
\newblock


\bibitem[\protect\citeauthoryear{Hirshman, Huang, and Macke}{Hirshman
  et~al\mbox{.}}{2013}]%
        {Hirshman2013}
\bibfield{author}{\bibinfo{person}{Jason Hirshman}, \bibinfo{person}{Yifei
  Huang}, {and} \bibinfo{person}{Stephen Macke}.}
  \bibinfo{year}{2013}\natexlab{}.
\newblock \showarticletitle{Unsupervised approaches to detecting anomalous
  behavior in the bitcoin transaction network}.
\newblock \bibinfo{journal}{\emph{3rd ed. Technical report, Stanford
  University}} (\bibinfo{year}{2013}).
\newblock


\bibitem[\protect\citeauthoryear{Hu, Seneviratne, Thilakarathna, Fukuda, and
  Seneviratne}{Hu et~al\mbox{.}}{2019}]%
        {Hu2019}
\bibfield{author}{\bibinfo{person}{Yining Hu}, \bibinfo{person}{Suranga
  Seneviratne}, \bibinfo{person}{Kanchana Thilakarathna},
  \bibinfo{person}{Kensuke Fukuda}, {and} \bibinfo{person}{Aruna Seneviratne}.}
  \bibinfo{year}{2019}\natexlab{}.
\newblock \showarticletitle{Characterizing and Detecting Money Laundering
  Activities on the Bitcoin Network}.
\newblock \bibinfo{journal}{\emph{arXiv preprint arXiv:1912.12060}}
  (\bibinfo{year}{2019}).
\newblock


\bibitem[\protect\citeauthoryear{Larik and Haider}{Larik and Haider}{2011}]%
        {Larik2011}
\bibfield{author}{\bibinfo{person}{Asma~S Larik} {and} \bibinfo{person}{Sajjad
  Haider}.} \bibinfo{year}{2011}\natexlab{}.
\newblock \showarticletitle{Clustering based anomalous transaction reporting}.
\newblock \bibinfo{journal}{\emph{Procedia Computer Science}}
  \bibinfo{volume}{3} (\bibinfo{year}{2011}), \bibinfo{pages}{606--610}.
\newblock


\bibitem[\protect\citeauthoryear{Laws and Sch{\"a}tze}{Laws and
  Sch{\"a}tze}{2008}]%
        {Laws2008}
\bibfield{author}{\bibinfo{person}{Florian Laws} {and} \bibinfo{person}{Hinrich
  Sch{\"a}tze}.} \bibinfo{year}{2008}\natexlab{}.
\newblock \showarticletitle{Stopping criteria for active learning of named
  entity recognition}. In \bibinfo{booktitle}{\emph{Proceedings of the 22nd
  International Conference on Computational Linguistics-Volume 1}}. Association
  for Computational Linguistics, \bibinfo{pages}{465--472}.
\newblock


\bibitem[\protect\citeauthoryear{Lewis and Catlett}{Lewis and Catlett}{1994}]%
        {Lewis1994}
\bibfield{author}{\bibinfo{person}{David~D Lewis} {and} \bibinfo{person}{Jason
  Catlett}.} \bibinfo{year}{1994}\natexlab{}.
\newblock \showarticletitle{Heterogeneous uncertainty sampling for supervised
  learning}.
\newblock In \bibinfo{booktitle}{\emph{Machine learning proceedings 1994}}.
  \bibinfo{publisher}{Elsevier}, \bibinfo{pages}{148--156}.
\newblock


\bibitem[\protect\citeauthoryear{Li, Cao, Qiu, Zhao, and Zheng}{Li
  et~al\mbox{.}}{2017}]%
        {Li2017}
\bibfield{author}{\bibinfo{person}{Xurui Li}, \bibinfo{person}{Xiang Cao},
  \bibinfo{person}{Xuetao Qiu}, \bibinfo{person}{Jintao Zhao}, {and}
  \bibinfo{person}{Jianbin Zheng}.} \bibinfo{year}{2017}\natexlab{}.
\newblock \showarticletitle{Intelligent anti-money laundering solution based
  upon novel community detection in massive transaction networks on spark}. In
  \bibinfo{booktitle}{\emph{2017 fifth international conference on advanced
  cloud and big data (CBD)}}. IEEE, \bibinfo{pages}{176--181}.
\newblock


\bibitem[\protect\citeauthoryear{Liu and Zhang}{Liu and Zhang}{2010}]%
        {Liu2010}
\bibfield{author}{\bibinfo{person}{Xuan Liu} {and} \bibinfo{person}{Pengzhu
  Zhang}.} \bibinfo{year}{2010}\natexlab{}.
\newblock \showarticletitle{A scan statistics based Suspicious transactions
  detection model for Anti-Money Laundering (AML) in financial institutions}.
  In \bibinfo{booktitle}{\emph{2010 International Conference on Multimedia
  Communications}}. IEEE, \bibinfo{pages}{210--213}.
\newblock


\bibitem[\protect\citeauthoryear{Liu, Zhang, and Zeng}{Liu
  et~al\mbox{.}}{2008}]%
        {Liu2008}
\bibfield{author}{\bibinfo{person}{Xuan Liu}, \bibinfo{person}{Pengzhu Zhang},
  {and} \bibinfo{person}{Dajun Zeng}.} \bibinfo{year}{2008}\natexlab{}.
\newblock \showarticletitle{Sequence matching for suspicious activity detection
  in anti-money laundering}. In \bibinfo{booktitle}{\emph{International
  Conference on Intelligence and Security Informatics}}. Springer,
  \bibinfo{pages}{50--61}.
\newblock


\bibitem[\protect\citeauthoryear{McInnes, Healy, and Melville}{McInnes
  et~al\mbox{.}}{2018}]%
        {McInnes2018}
\bibfield{author}{\bibinfo{person}{Leland McInnes}, \bibinfo{person}{John
  Healy}, {and} \bibinfo{person}{James Melville}.}
  \bibinfo{year}{2018}\natexlab{}.
\newblock \showarticletitle{{UMAP: Uniform Manifold Approximation and
  Projection for Dimension Reduction}}.
\newblock  (\bibinfo{year}{2018}).
\newblock


\bibitem[\protect\citeauthoryear{Monamo, Marivate, and Twala}{Monamo
  et~al\mbox{.}}{2016a}]%
        {Monamo2016}
\bibfield{author}{\bibinfo{person}{Patrick Monamo}, \bibinfo{person}{Vukosi
  Marivate}, {and} \bibinfo{person}{Bheki Twala}.}
  \bibinfo{year}{2016}\natexlab{a}.
\newblock \showarticletitle{Unsupervised learning for robust Bitcoin fraud
  detection}. In \bibinfo{booktitle}{\emph{2016 Information Security for South
  Africa (ISSA)}}. IEEE, \bibinfo{pages}{129--134}.
\newblock


\bibitem[\protect\citeauthoryear{Monamo, Marivate, and Twala}{Monamo
  et~al\mbox{.}}{2016b}]%
        {Monamo2016_multifaceted}
\bibfield{author}{\bibinfo{person}{Patrick~M Monamo}, \bibinfo{person}{Vukosi
  Marivate}, {and} \bibinfo{person}{Bhesipho Twala}.}
  \bibinfo{year}{2016}\natexlab{b}.
\newblock \showarticletitle{A multifaceted approach to bitcoin fraud detection:
  Global and local outliers}. In \bibinfo{booktitle}{\emph{2016 15th IEEE
  International Conference on Machine Learning and Applications (ICMLA)}}.
  IEEE, \bibinfo{pages}{188--194}.
\newblock


\bibitem[\protect\citeauthoryear{Monroe}{Monroe}{2020}]%
        {monroe_ousted_2020}
\bibfield{author}{\bibinfo{person}{Brian Monroe}.}
  \bibinfo{year}{2020}\natexlab{}.
\newblock \bibinfo{title}{Ousted {Danske} {Bank} {CEO} faces nearly \$400
  million lawsuit tied to historic money laundering scandal}.
\newblock
\newblock
\urldef\tempurl%
\url{https://www.acfcs.org/ousted-danske-bank-ceo-faces-nearly-400-million-lawsuit-tied-to-historic-money-laundering-scandal/}
\showURL{%
\tempurl}
\newblock
\shownote{Library Catalog: www.acfcs.org Section: Uncategorized.}


\bibitem[\protect\citeauthoryear{Network}{Network}{2019}]%
        {financial_crimes_enforcement_network_application_2019}
\bibfield{author}{\bibinfo{person}{Financial Crimes~Enforcement Network}.}
  \bibinfo{year}{2019}\natexlab{}.
\newblock \bibinfo{title}{Application of {FinCEN}’s {Regulations} to
  {Certain} {Business} {Models} {Involving} {Convertible} {Virtual}
  {Currencies} {\textbar} {FinCEN}.gov}.
\newblock
\newblock
\urldef\tempurl%
\url{https://www.fincen.gov/resources/statutes-regulations/guidance/application-fincens-regulations-certain-business-models}
\showURL{%
\tempurl}


\bibitem[\protect\citeauthoryear{Noonan, Palma, and Shubber}{Noonan
  et~al\mbox{.}}{2020}]%
        {1mdb_malaysia_2020}
\bibfield{author}{\bibinfo{person}{Laura Noonan}, \bibinfo{person}{Stefania
  Palma}, {and} \bibinfo{person}{Kadhim Shubber}.}
  \bibinfo{year}{2020}\natexlab{}.
\newblock \showarticletitle{The 1MDB scandal: what does it mean for Goldman
  Sachs?}
\newblock \bibinfo{journal}{\emph{Financial Times}} (\bibinfo{date}{Jan}
  \bibinfo{year}{2020}).
\newblock
\urldef\tempurl%
\url{https://www.ft.com/content/3f161eda-3306-11ea-9703-eea0cae3f0de}
\showURL{%
\tempurl}


\bibitem[\protect\citeauthoryear{Pedregosa, Varoquaux, Gramfort, Michel,
  Thirion, Grisel, Blondel, Prettenhofer, Weiss, Dubourg, Vanderplas, Passos,
  Cournapeau, Brucher, Perrot, and Duchesnay}{Pedregosa et~al\mbox{.}}{2011}]%
        {scikit-learn}
\bibfield{author}{\bibinfo{person}{F. Pedregosa}, \bibinfo{person}{G.
  Varoquaux}, \bibinfo{person}{A. Gramfort}, \bibinfo{person}{V. Michel},
  \bibinfo{person}{B. Thirion}, \bibinfo{person}{O. Grisel},
  \bibinfo{person}{M. Blondel}, \bibinfo{person}{P. Prettenhofer},
  \bibinfo{person}{R. Weiss}, \bibinfo{person}{V. Dubourg}, \bibinfo{person}{J.
  Vanderplas}, \bibinfo{person}{A. Passos}, \bibinfo{person}{D. Cournapeau},
  \bibinfo{person}{M. Brucher}, \bibinfo{person}{M. Perrot}, {and}
  \bibinfo{person}{E. Duchesnay}.} \bibinfo{year}{2011}\natexlab{}.
\newblock \showarticletitle{Scikit-learn: Machine Learning in {P}ython}.
\newblock \bibinfo{journal}{\emph{Journal of Machine Learning Research}}
  \bibinfo{volume}{12} (\bibinfo{year}{2011}), \bibinfo{pages}{2825--2830}.
\newblock


\bibitem[\protect\citeauthoryear{Pham and Lee}{Pham and Lee}{2016a}]%
        {Pham2016_unsupervised}
\bibfield{author}{\bibinfo{person}{Thai Pham} {and} \bibinfo{person}{Steven
  Lee}.} \bibinfo{year}{2016}\natexlab{a}.
\newblock \showarticletitle{Anomaly detection in bitcoin network using
  unsupervised learning methods}.
\newblock \bibinfo{journal}{\emph{arXiv preprint arXiv:1611.03941}}
  (\bibinfo{year}{2016}).
\newblock


\bibitem[\protect\citeauthoryear{Pham and Lee}{Pham and Lee}{2016b}]%
        {Pham2016_networkperspective}
\bibfield{author}{\bibinfo{person}{Thai Pham} {and} \bibinfo{person}{Steven
  Lee}.} \bibinfo{year}{2016}\natexlab{b}.
\newblock \showarticletitle{Anomaly detection in the bitcoin system-a network
  perspective}.
\newblock \bibinfo{journal}{\emph{arXiv preprint arXiv:1611.03942}}
  (\bibinfo{year}{2016}).
\newblock


\bibitem[\protect\citeauthoryear{Raza and Haider}{Raza and Haider}{2011}]%
        {Raza2011}
\bibfield{author}{\bibinfo{person}{Saleha Raza} {and} \bibinfo{person}{Sajjad
  Haider}.} \bibinfo{year}{2011}\natexlab{}.
\newblock \showarticletitle{Suspicious activity reporting using dynamic
  bayesian networks}.
\newblock \bibinfo{journal}{\emph{Procedia Computer Science}}
  \bibinfo{volume}{3} (\bibinfo{year}{2011}), \bibinfo{pages}{987--991}.
\newblock


\bibitem[\protect\citeauthoryear{Settles}{Settles}{2009}]%
        {Settles2009}
\bibfield{author}{\bibinfo{person}{Burr Settles}.}
  \bibinfo{year}{2009}\natexlab{}.
\newblock \bibinfo{booktitle}{\emph{Active learning literature survey}}.
\newblock \bibinfo{type}{{T}echnical {R}eport}.
  \bibinfo{institution}{University of Wisconsin-Madison Department of Computer
  Sciences}.
\newblock


\bibitem[\protect\citeauthoryear{Settles, Craven, and Ray}{Settles
  et~al\mbox{.}}{2008}]%
        {Settles2008}
\bibfield{author}{\bibinfo{person}{Burr Settles}, \bibinfo{person}{Mark
  Craven}, {and} \bibinfo{person}{Soumya Ray}.}
  \bibinfo{year}{2008}\natexlab{}.
\newblock \showarticletitle{Multiple-instance active learning}. In
  \bibinfo{booktitle}{\emph{Advances in neural information processing
  systems}}, Vol.~\bibinfo{volume}{20}. \bibinfo{pages}{1289--1296}.
\newblock


\bibitem[\protect\citeauthoryear{Stokes, Platt, Kravis, and Shilman}{Stokes
  et~al\mbox{.}}{2008}]%
        {Stokes2008}
\bibfield{author}{\bibinfo{person}{Jack~W Stokes}, \bibinfo{person}{John
  Platt}, \bibinfo{person}{Joseph Kravis}, {and} \bibinfo{person}{Michael
  Shilman}.} \bibinfo{year}{2008}\natexlab{}.
\newblock \showarticletitle{Aladin: Active learning of anomalies to detect
  intrusions}.
\newblock  (\bibinfo{year}{2008}).
\newblock


\bibitem[\protect\citeauthoryear{Sudjianto, Nair, Yuan, Zhang, Kern, and
  Cela-D{\'\i}az}{Sudjianto et~al\mbox{.}}{2010}]%
        {sudjianto2010statistical}
\bibfield{author}{\bibinfo{person}{Agus Sudjianto}, \bibinfo{person}{Sheela
  Nair}, \bibinfo{person}{Ming Yuan}, \bibinfo{person}{Aijun Zhang},
  \bibinfo{person}{Daniel Kern}, {and} \bibinfo{person}{Fernando
  Cela-D{\'\i}az}.} \bibinfo{year}{2010}\natexlab{}.
\newblock \showarticletitle{Statistical methods for fighting financial crimes}.
\newblock \bibinfo{journal}{\emph{Technometrics}} \bibinfo{volume}{52},
  \bibinfo{number}{1} (\bibinfo{year}{2010}), \bibinfo{pages}{5--19}.
\newblock


\bibitem[\protect\citeauthoryear{Tang and Yin}{Tang and Yin}{2005}]%
        {Tang2005}
\bibfield{author}{\bibinfo{person}{Jun Tang} {and} \bibinfo{person}{Jian Yin}.}
  \bibinfo{year}{2005}\natexlab{}.
\newblock \showarticletitle{Developing an intelligent data discriminating
  system of anti-money laundering based on SVM}. In
  \bibinfo{booktitle}{\emph{2005 International conference on machine learning
  and cybernetics}}, Vol.~\bibinfo{volume}{6}. IEEE,
  \bibinfo{pages}{3453--3457}.
\newblock


\bibitem[\protect\citeauthoryear{Union}{Union}{2018}]%
        {european_union_directive_2018}
\bibfield{author}{\bibinfo{person}{European Union}.}
  \bibinfo{year}{2018}\natexlab{}.
\newblock \showarticletitle{Directive ({EU}) 2018/843 of the {European}
  {Parliament} and of the {Council} of 30 {May} 2018 amending {Directive}
  ({EU}) 2015/849 on the prevention of the use of the financial system for the
  purposes of money laundering or terrorist financing, and amending
  {Directives} 2009/138/{EC} and 2013/36/{EU}}.
\newblock \bibinfo{journal}{\emph{Official Journal of the European Union}}
  \bibinfo{volume}{L 156} (\bibinfo{date}{June} \bibinfo{year}{2018}),
  \bibinfo{pages}{43--74}.
\newblock


\bibitem[\protect\citeauthoryear{Wang and Dong}{Wang and Dong}{2009}]%
        {Wang2009}
\bibfield{author}{\bibinfo{person}{Xingqi Wang} {and} \bibinfo{person}{Guang
  Dong}.} \bibinfo{year}{2009}\natexlab{}.
\newblock \showarticletitle{Research on money laundering detection based on
  improved minimum spanning tree clustering and its application}. In
  \bibinfo{booktitle}{\emph{2009 Second international symposium on knowledge
  acquisition and modeling}}, Vol.~\bibinfo{volume}{2}. IEEE,
  \bibinfo{pages}{62--64}.
\newblock


\bibitem[\protect\citeauthoryear{Wang, Wang, Gao, and Xu}{Wang
  et~al\mbox{.}}{2008}]%
        {Wang2008_intelligent}
\bibfield{author}{\bibinfo{person}{Yingfeng Wang}, \bibinfo{person}{Huaiquing
  Wang}, \bibinfo{person}{Caddie Shi~Jia Gao}, {and} \bibinfo{person}{Dongming
  Xu}.} \bibinfo{year}{2008}\natexlab{}.
\newblock \showarticletitle{Intelligent money laundering monitoring and
  detecting system}. In \bibinfo{booktitle}{\emph{European, Mediterranean and
  Middle Eastern Conference on Information Systems 2008}}. Brunel University,
  \bibinfo{pages}{1--11}.
\newblock


\bibitem[\protect\citeauthoryear{Weber, Domeniconi, Chen, Weidele, Bellei,
  Robinson, and Leiserson}{Weber et~al\mbox{.}}{2019}]%
        {Weber2019}
\bibfield{author}{\bibinfo{person}{Mark Weber}, \bibinfo{person}{Giacomo
  Domeniconi}, \bibinfo{person}{Jie Chen}, \bibinfo{person}{Daniel Karl~I
  Weidele}, \bibinfo{person}{Claudio Bellei}, \bibinfo{person}{Tom Robinson},
  {and} \bibinfo{person}{Charles~E Leiserson}.}
  \bibinfo{year}{2019}\natexlab{}.
\newblock \showarticletitle{Anti-money laundering in bitcoin: Experimenting
  with graph convolutional networks for financial forensics}.
\newblock \bibinfo{journal}{\emph{arXiv preprint arXiv:1908.02591}}
  (\bibinfo{year}{2019}).
\newblock


\bibitem[\protect\citeauthoryear{Wu, Pan, Chen, Long, Zhang, and Philip}{Wu
  et~al\mbox{.}}{2020}]%
        {Wu2020}
\bibfield{author}{\bibinfo{person}{Zonghan Wu}, \bibinfo{person}{Shirui Pan},
  \bibinfo{person}{Fengwen Chen}, \bibinfo{person}{Guodong Long},
  \bibinfo{person}{Chengqi Zhang}, {and} \bibinfo{person}{S~Yu Philip}.}
  \bibinfo{year}{2020}\natexlab{}.
\newblock \showarticletitle{A comprehensive survey on graph neural networks}.
\newblock \bibinfo{journal}{\emph{IEEE Transactions on Neural Networks and
  Learning Systems}} (\bibinfo{year}{2020}).
\newblock


\bibitem[\protect\citeauthoryear{Zhao, Nasrullah, and Li}{Zhao
  et~al\mbox{.}}{2019}]%
        {zhao2019pyod}
\bibfield{author}{\bibinfo{person}{Yue Zhao}, \bibinfo{person}{Zain Nasrullah},
  {and} \bibinfo{person}{Zheng Li}.} \bibinfo{year}{2019}\natexlab{}.
\newblock \showarticletitle{PyOD: A Python Toolbox for Scalable Outlier
  Detection}.
\newblock \bibinfo{journal}{\emph{Journal of Machine Learning Research}}
  \bibinfo{volume}{20}, \bibinfo{number}{96} (\bibinfo{year}{2019}),
  \bibinfo{pages}{1--7}.
\newblock


\end{thebibliography}





\end{document}